\documentclass[11pt,a4paper]{article}
\usepackage[hyperref]{acl2021}
\usepackage{times}
\usepackage{latexsym}

\usepackage{amsfonts} 
\usepackage{microtype}      
\usepackage{amsmath,amssymb}
\usepackage{multirow}       
\usepackage{subfigure}
\usepackage{graphicx}
\usepackage{booktabs}
\usepackage{enumitem} 

\usepackage{microtype}

\aclfinalcopy 

\title{Unified Dual-view Cognitive Model for Interpretable Claim Verification}

\author{Lianwei Wu, Yuan Rao, Yuqian Lan, Ling Sun, Zhaoyin Qi \\
  Xi'an Key Lab. of Social Intelligence and Complexity Data Processing, \\ School of Software Engineering, Xi'an Jiaotong University, China \\
  \textls[-30]{Shaanxi Joint Key Laboratory for Artifact Intelligence\! (Sub-Lab of Xi'an Jiaotong University), China} \\
  Research Institute of Xi'an Jiaotong University, Shenzhen, China \\
  {\tt {\small \{stayhungry,yuqian\_lan\_xjtu,sunling,daqige123\}@stu.xjtu.edu.cn}} \\ {\small {\tt raoyuan@mail.xjtu.edu.cn}} \\}

\date{}

\begin{document}
\maketitle
\begin{abstract}
 \textls[-30]{Recent studies constructing direct interactions between the claim and each single user response to capture evidence have shown remarkable success in interpretable claim verification. Owing to different single responses convey different cognition of individual users, the captured evidence belongs to the perspective of individual cognition. However, individuals' cognition of social things is not always able to truly reflect the objective. There may be one-sided or biased semantics in their opinions on a claim. The captured evidence correspondingly contains some unobjective and biased information. In this paper, we propose a Dual-view model based on the views of Collective and Individual Cognition (CICD) for interpretable claim verification. For collective cognition, we not only capture the word-level semantics based on individual users, but also focus on sentence-level semantics (i.e., the overall responses) among all users to generate global evidence. For individual cognition, we select the top-$k$ articles with high degree of difference and interact with the claim to explore the local key evidence fragments. To weaken the bias of individual cognition-view evidence, we devise an inconsistent loss to suppress the divergence between global and local evidence for strengthening the consistent shared evidence between the both. Experiments on three benchmark datasets confirm the effectiveness of CICD.}
\end{abstract}

\section{Introduction}
\textls[-35]{The problem of claim credibility has seriously affected the media ecosystem. Research \cite{allen2020evaluating} illustrates that the prevalence of `fake news' has decreased trust in public institutions, and undermined democracy. Meanwhile, `massive infodemic' during COVID-19 has taken a great toll on health-care systems and lives \cite{nic2020coronavirus}. Therefore, how to verify the claims spread in networks has become a crucial issue.}

\textls[-20]{Current approaches on claim verification could be divided into two categories: 1) The first category relies on traditional machine learning and deep learning methods to capture semantics \cite{yang2019unsupervised}, sentiments \cite{ajao2019sentiment}, writing styles \cite{przybyla2020capturing}, and stances \cite{kumar2019tree} from claim content, and meta-data features, such as user profiles \cite{shu2019role,wu2020discovering} for verification. Such approaches could improve verification performance, but they are hard to make reasonable explanations for the verified results, i.e., where false claims go wrong; 2) To tackle this issue, many researchers further focus on interpretable claim verification (the second category) by establishing interactive models between claims and each individual relevant article (or comment) to explore coherent \cite{ma2019sentence,wu2021evidence}, similar \cite{nie2019combining,wu2020dtca}, or conflicting \cite{zhou2020safe} semantics as evidence for verifying the false parts of claims.}

\begin{figure}
	\centering
	\includegraphics[width=0.45\textwidth]{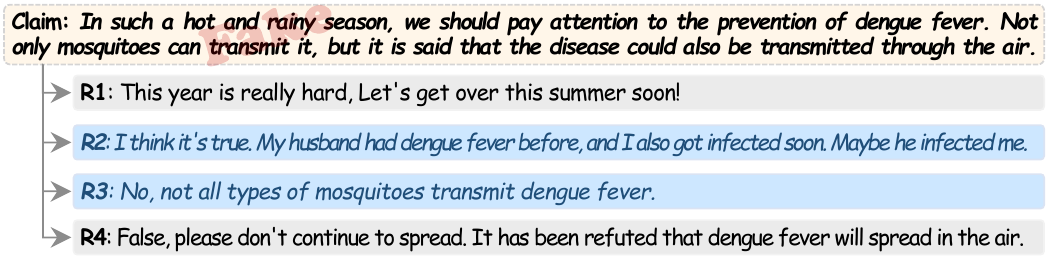}
	\caption{Users' individual cognition of a false claim.}
	\label{fig1example}
\end{figure}

\textls[-15]{In interpretable claim verification, the majority of models construct interactions between claims and each single user response (i.e., a comment or a relevant article) to capture evidence, which could effectively learn some of errant aspects of false claims. Due to different single responses reflect the cognition of different individual users, the evidence captured by these models is usually confined to individual cognition. However, individuals' cognition of social things is not always able to truly reflect the objective \cite{Greenwald1998MeasuringID,Boogert2018MeasuringAU}. Owing to individuals are affected by factors such as emotional tendency \cite{Ji2019SpontaneousCI}, traditional beliefs \cite{Willard2017SpiritualBN}, and selectively capturing information \cite{Hoffman2018AnID}, there are considerable differences in cognition of different individuals, and they are prone to cognitive bias, like primacy effect \cite{Troyer2011} and halo effect \cite{Goldstein2011halo}, there may be one-sided or biased semantics in their expressed opinions. Thus, the captured evidence also correspondingly contains some unobjective and biased evidence fragments, deteriorating task performance. For instance, as shown in Figure \ref{fig1example}, facing a claim to be verified, different individual users (here, users are the normal users on social media, not journalists or professionals) have different reactions. R2 (i.e., response 2 or relevant article 2) and R3 released by users contain unreliable and biased information perceived by their individuals, which may lead to some misleading information being captured as evidence by existing interactive models. Therefore, how to explore users' collective cognition on claims is a major challenge for interpretable claim verification.}

\textls[-25]{To address the deficiencies, we propose a unified \textbf{D}ual-view model based on \textbf{C}ollective and \textbf{I}ndividual \textbf{C}ognition (CICD) for interpretable claim verification, which focuses on discovering global evidence and local key evidence, respectively, and then strengthens the consistent shared evidence between the both. Specifically, to explore users' collective cognition to capture global evidence, we design Collective cognition view-based Encoder-Decoder module (CED). CED develops claim-guided encoder that not only learns word-level semantics based on individual user, but also captures sentence-level semantics (i.e., the overall opinions) among all users. Here, a relevant article (a response) released by an individual user is usually a sentence sequence, so all sentence-level semantics convey the overall opinions of all users. Then, CED develops hierarchical attention decoder to generate global evidence by adjusting weights of word-level and sentence-level semantics. To further acquire the local key evidence based on individual cognition, we develop Individual cognition view-based Selected Interaction module (ISI) to screen representative top-$k$ articles with high difference and interact with the claim to gain local key evidence fragments. To weaken the bias of individual cognition view and strengthen the consistent shared evidence between global and local evidence, we project inconsistent loss to suppress the divergence. Experimental results not only reveal the effectiveness of CICD but also provide its interpretability. Our contributions are summarized:}

\begin{itemize}[leftmargin=*]
\item \textls[-10]{A novel framework integrating interdisciplinary knowledge on interpretable claim verification is explored, which discovers global and local evidence from the perspectives of collective and individual cognition to interpret verified results.}
\item \textls[-20]{Proposed CED captures word-level (individual) and sentence-level (holistic) opinions, and reasonably adjusts the proportion between them, which generates global evidence of the view of all users.}
\item \textls[-15]{Experiments on three competitive datasets demonstrate that CICD achieves better performance than other strong baselines.}
\end{itemize}

\vspace{-0.1cm}
\section{Related Work}
\vspace{-0.0cm}


\textls[-20]{Automatic verification approaches rely on neural networks to extract content-based features, like semantics \cite{popat2018declare,wu2019different}, sentiments \cite{nguyen2020fang}, writing styles \cite{przybyla2020capturing}, etc., and metadata-based features, like user profiles-based \cite{kumar2019tree}, comment-based \cite{bovet2019influence}, etc., for verification. These methods could improve the accuracy of claim verification, but they are lack of interpretability for the verified results. To tackle this, interpretable claim verification has received great attention. Its basic principle is to obtain queried, corrected, and rumor-refuted semantics from the articles (or comments) related to claims to interpret the credibility of claims. At present, the methods for this task generally focus on direct interactions between claims and relevant articles to identify their matching degree \cite{nie2019combining}, consistency \cite{ma2019sentence}, implication \cite{liu2019trust}, conflict \cite{wu2020evidence}, etc., to learn practical evidence. For instances, HAN \cite{ma2019sentence} and EHIAN \cite{wu2020evidence} learned implication relationships between claims and relevant articles to capture semantic conflicts as evidence, which reflected a certain interpretability. However, since all relevant articles are involved, the captured conflicts may be affected by some low-quality articles with noisy semantics, easily resulting in the invalidation of the evidence. In our model, we design ISI module to screen all relevant articles to capture the valuable representative articles with differential semantics, so as to learn local key evidence fragments. In addition, some methods, such as GEAR \cite{zhou2019gear} and KGAT \cite{liu2020fine}, relied on graph-based networks to conduct semantic aggregation and reasoning on relevant articles, so as to capture global evidence. Nevertheless, these models treat an entire article (at the sentence level) as a node and ignore the importance of word-level semantics in each article. To overcome these defects, our model constructs a hierarchical attention decoder to fuse sentence-level and word-level semantics for finely-grained generating global evidence.}

\begin{figure*}
	\centering
	\includegraphics[width=0.85\textwidth]{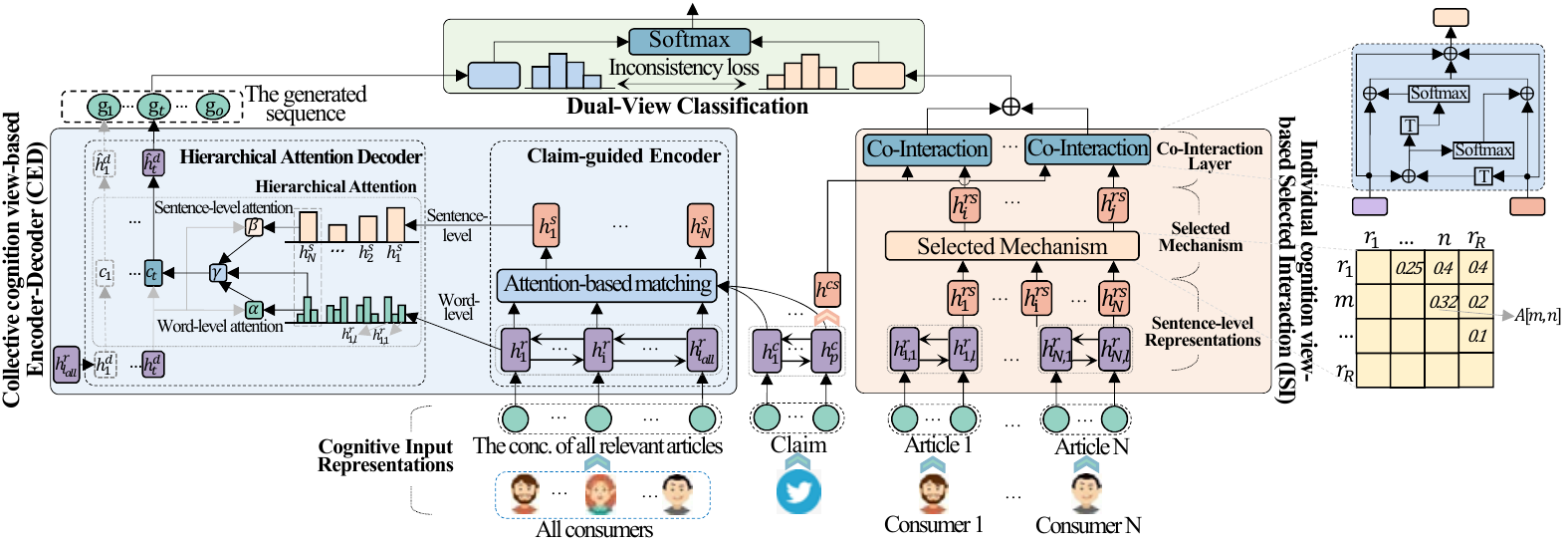}
	\caption{The general architecture of CICD. The model consists of: 1) CED for generating global evidence; 2) ISI for exploring local key evidence fragments; and 3) Dual-view classification module leveraging inconsistency loss to promote the learning of shared available evidence between global and local evidence.}
	\label{fig2model}
\end{figure*}

\section{The Proposed Approach}
In this section, we introduce the details of CICD as illustrated in Figure \ref{fig2model}.

\textbf{Inputs and Outputs}
\textls[-20]{For cognitive input representations, the inputs of CED are a claim sequence and the concatenation of its all relevant articles with the number of $N$, while the inputs of ISI are a claim sequence and each relevant article. Given any a sequence of length $l$ words ${\rm \textbf{X}}\!=\!\{x_1, x_2, ..., x_l\}$, where each word $x_i \!\in \! \mathbb{R}^d$ is a $d$-dimensional vector obtained by pre-trained BERT model \cite{devlin2019bert}. Particularly, the length of each sequence in relevant articles is $l$ and that of the claim sequence is $p$. Thus, we obtain the representations of the $i$-th relevant article and the claim as ${\rm \textbf{X}}_i^r \!\in\! \mathbb{R}^{l\times d}$, ${\rm \textbf{X}}^c \!\in\! \mathbb{R}^{p\times d}$, respectively. For the outputs of the model, the outputs of CED are the generated global evidence sequence of length $o$ words ${\rm \textbf{G}}\!=\!\{g_1, g_2, ..., g_o\}$, where $g_t$ is the representation of the $t$-th generated word and $o$ is the length of ${\rm \textbf{G}}$. The outputs of ISI are the integrated vector of top-$k$ local key evidence fragments ${\rm \textbf{I}}\!=\![{\rm \textbf{I}}_1; {\rm \textbf{I}}_2,; ...; {\rm \textbf{I}}_k]$, where $;$ is the concatenation operation.}

\subsection{Collective Cognition View-based Encoder-Decoder (CED)}
\textls[0]{To explore users' collective cognition on claims, we first rely on claim-guided encoder to capture word-level and sentence-level semantics from all relevant articles, and then adjust the proportion between the both by hierarchical attention decoder to generate global evidence.}
\subsubsection{Claim-guided Encoder}
\label{sec32claim}
The claim-guided encoder module involves a sequence encoding layer and a matching layer.

\textbf{Sequence Encoding Layer } \textls[-15]{We rely on BiLSTMs to encode all relevant articles and the claim for their contextual representations. We utilize the produced hidden states ${\rm \textbf{H}}^{r} \!=\! \{{\rm h}_1^r, {\rm h}_2^r, ..., {\rm h}_{l_{all}}^r\}$ (where $l_{all}$  means the total length of all articles) and ${\rm \textbf{H}}^c\!=\!\{{\rm h}_1^c, {\rm h}_2^c, ..., {\rm h}_p^c\}$ to denote the contextual representations of relevant articles and the claim, respectively, where ${\rm h}_i$ (i.e., ${\rm h}_i^r$ or ${\rm h}_i^c$) is defined as follows:}
\begin{equation}\label{eq1}
\setlength{\abovedisplayskip}{0pt}
\setlength{\belowdisplayskip}{0pt}
{\rm h}_i = [\overrightarrow{{\rm h}_i}; \overleftarrow{{\rm h}_i}]
\end{equation}
\textls[-5]{where $\overrightarrow{{\rm h}_i} \!\in\! \mathbb{R}^{d_h}$ and $\overleftarrow{{\rm h}_i}\!\in\! \mathbb{R}^{d_h}$ are the $i$-th hidden state of the forward and backward LSTMs for the word $x_i$ respectively. $;$ is concatenation operation.}

\textls[-30]{\textbf{Attention-based Matching Layer } is engaged to aggregate the relevant information from the claim for each word within the context of relevant articles. The aggregation operation $a_i\!\!=\!\!{\rm attn}({\rm h}_i^r, {\rm \textbf{H}}^c)$ is as follows:}
\begin{align}
\setlength{\abovedisplayskip}{0pt}
\setlength{\belowdisplayskip}{0pt}
a_i &=\sum_{j=1}^{k_c}\alpha_{i,j} {\rm h}_j^c  \label{eq2} \\
\alpha_{i,j}\!&=\!{\rm exp}(s_{i,j})/\sum_{k=1}^{p}{\rm exp}(s_{i,k})  \label{eq3}\\
s_{i,j}&=({\rm h}_i^r)^\top {\rm \textbf{W}}_1{\rm h}_j^c  \label{eq4}
\end{align}
\textls[-15]{where $a_i$ is the aggregated vector for the $i$-th word of the articles. $\alpha_{i,j}$ is the normalized attention score between ${\rm h}_i^r$ and ${\rm h}_j^c$.}

\textls[-20]{Here, the purpose of adopting claim to guide the encoding of relevant articles includes two perspectives: 1) Strengthening the focus of consistent semantics associated with the claim in relevant articles, i.e., exploring how relevant articles evaluate the claim; and 2) Making the encoding semantics purer. We observe that there are some advertisements or useless information in relevant articles. This way is able to effectively filter the noise irrelevant to the claim from relevant articles, and consolidates the generation of relevant semantics in the decoder module.}

\textls[-25]{Furthermore, we output the hidden state corresponding to the last word encoded by each relevant article to form consistent sentence-level representations, where ${\rm h}_i^s$ represents sentence-level representations of the $i$-th relevant article. Particularly, we apply word-level representations ${\rm \textbf{H}}^{r} \!\!=\!\! \{{\rm h}_1^r, {\rm h}_2^r, ..., {\rm h}_{l_{all}}^r\}$ (which can also be represented in the form of different relevant articles, i.e., ${\rm \textbf{H}}^{r} \!\!=\!\! \{{\rm h}_{1,1}^r, {\rm h}_{1,2}^r, ..., {\rm h}_{N,l}^r\}$, where $l_{all}\!\!=\!\!N \!\!\times\!\! l$) and sentence-level representations ${\rm \textbf{H}}^{rs} \!\!=\!\! \{{\rm h}_1^s, {\rm h}_2^s, ..., {\rm h}_N^s\}$ as memory bank for decoder generation.}

\subsubsection{Hierarchical Attention Decoder}
\textls[-20]{To capture the collective cognition-view evidence from relevant articles, we devise hierarchical attention decoder to consider the consistent semantics with different granularity of relevant articles to generate global evidence. Specifically, we employ uni-directional LSTM as the decoder, and at each decoding time-step, we calculate in parallel both sentence-level attention weight $\beta$ and word-level $\alpha$ by:}
\begin{equation}\label{eq6}
\setlength{\abovedisplayskip}{2pt}
\setlength{\belowdisplayskip}{0pt}
\beta_i =(h_i^s)^\top {\rm \textbf{W}}_2 h_t^d  \quad \quad
\alpha_{i,j} =(h_{i,j}^r)^\top {\rm \textbf{W}}_3 h_t^d
\end{equation}
\begin{equation}\label{eq7}
\setlength{\abovedisplayskip}{2pt}
\setlength{\belowdisplayskip}{0pt}
\gamma_{i,j}=\frac{\alpha_{i,j}\beta_i}{\sum_{i,j}\alpha_{i,j}\beta_i}
\end{equation}
\textls[-15]{where ${\rm h}_t^d$ is the hidden state of the decoder at the $t$-th time-step, ${\rm \textbf{W}}_2$ and ${\rm \textbf{W}}_3$ are trainable parameters. The word-level attention ascertains how to distribute the attention over words in each sentence (each article), which could learn salient evidence segments in each article, while the sentence-level attention determines how much each article should contribute to the generation at current time-step, which could capture potential global semantics in all articles.}

Then the context vector ${\rm c}_t$ is derived as a combination of all word-level representations reweighted by combined attention $\gamma$:
\begin{equation}\label{eq8}
\setlength{\abovedisplayskip}{2pt}
\setlength{\belowdisplayskip}{2pt}
{\rm c}_t=\sum_{i,j}\gamma_{i,j}{\rm h}_i^r
\end{equation}

And the attentional vector is calculated as:
\begin{equation}\label{eq9}
\setlength{\abovedisplayskip}{2pt}
\setlength{\belowdisplayskip}{2pt}
\hat{{\rm h}}_t^d = {\rm tanh}({\rm \textbf{W}}_4[{\rm h}_t^d; {\rm c}_t])
\end{equation}

Finally, the predicted probability distribution over the vocabulary $V$ at the current step is:
\begin{equation}\label{eq10}
\setlength{\abovedisplayskip}{2pt}
\setlength{\belowdisplayskip}{2pt}
P_V = {\rm softmax}({\rm \textbf{W}}_V\hat{{\rm h}}_t^d + {\rm \textbf{b}}_V)
\end{equation}
where ${\rm \textbf{W}}_4$, ${\rm \textbf{W}}_V$, and ${\rm \textbf{b}}_V$ are trainable parameters.

We adopt ${\rm \textbf{G}}\!=\!\{g_1, g_2, ..., g_o\}$ to denote the generated sequence rich in global evidence.

\subsection{\textls[-0]{Individual Cognition View-based Selected Interaction (ISI)}}
\textls[-20]{To capture evidence fragments from individual cognition view, we design ISI module with the following layers: 1) Sentence-level representation for capturing high-level representations of relevant articles; 2) Selected mechanism for screening the representative top-$k$ relevant articles with degree of difference; and 3) Co-interaction layer for making the claim and the selected articles interact with each other to explore local key evidence fragments.}

\subsubsection{Sentence-level Representation}
We exploit BiLSTM to encode each relevant article and capture the output of the last hidden state as the sentence-level representation, where the encoding process is similar to sequence encoding layer in Section 3.2.1, where the sentence-level representation of the $i$-th article is ${\rm h}_i^{rs}$.

\subsubsection{Selected Mechanism}
To capture representative top-$k$ articles, we develop selected mechanism to calculate the difference between each articles and other articles in an automated manner. To do this, selected mechanism learns and optimizes an inter-sentential attention matrix ${\rm \textbf{A}} \!\in\! \mathbb{R}^{N \times N}$. The entry $(m, n)$ of ${\rm \textbf{A}}$ holds the difference between article $m$ and article $n$ ($1 \!\leq\! m,n \!\leq\! N$ and $m \!\neq\! n$) and is computed as:
\begin{equation}\label{eq11}
\setlength{\abovedisplayskip}{2pt}
\setlength{\belowdisplayskip}{2pt}
{\rm u}_m\!\!=\!\!\varphi({\rm \textbf{W}}_m {\rm h}_m^{rs} \!+\! {\rm \textbf{b}}_m) \quad
{\rm u}_n\!\!=\!\!\varphi({\rm \textbf{W}}_n {\rm h}_n^{rs} \!+\! {\rm \textbf{b}}_n)
\end{equation}
\begin{equation}\label{eq13}
\setlength{\abovedisplayskip}{2pt}
\setlength{\belowdisplayskip}{2pt}
{\rm \textbf{A}}[m, n] = \frac{{\rm exp}({\rm u}_m \odot {\rm u}_n)}{\sum_{i=1}^{N}{\rm exp}({\rm u}_i \odot {\rm u}_n)}
\end{equation}
\textls[-20]{where $\varphi$ is a activation function, ${\rm \textbf{W}}_m$ and ${\rm \textbf{W}}_n$ are weight matrix, ${\rm \textbf{b}}_m$ and ${\rm \textbf{b}}_n$ are biases, and $\odot$ denotes dot product operator. The larger the entry ${\rm \textbf{A}}[m, n]$ is, the higher the similar between article $m$ and article $n$ is. Thus, the smaller ${\rm \textbf{A}}[m, n]$ corresponds to article $m$ and $n$ contain more differential semantics, and finally we screen top-$k$ relevant articles with high difference for further downstream interaction.}

\subsubsection{Co-Interaction Layer}
\textls[-30]{This co-interaction layer aims to explore local key evidence fragments. Specifically, the layer enables the claim to focus on the $i$-th article to discover the specific evidence fragment, while the $i$-th article pays close attention to the claim to explore the possible false part of the claim. Finally, we combine the two interactions to constitute the individual key local evidence fragments.}
\begin{align}
\setlength{\abovedisplayskip}{2pt}
\setlength{\belowdisplayskip}{2pt}
{\rm \textbf{H}}_i^{rin}&= {\rm h}_i^{rs} +{\rm softmax}({\rm h}_i^{rs} (({\rm \textbf{H}}^{cs})^\top)){\rm \textbf{H}}^{cs}  \label{eq14} \\
{\rm \textbf{H}}^{cin}&={\rm \textbf{H}}^{cs}+{\rm softmax}({\rm \textbf{H}}^{cs} (({\rm h}_i^{rs})^\top)){\rm h}_i^{rs}  \label{eq15} \\
{\rm \textbf{I}}_i&=[{\rm \textbf{H}}_i^{rin}; {\rm \textbf{H}}^{cin}]  \label{eq16}
\end{align}
\textls[-15]{where ${\rm \textbf{H}}_i^{rin}$ is the evidence fragment of the $i$-th article, ${\rm \textbf{H}}^{cin}$ is the false part of the claim, and ${\rm \textbf{H}}^{cs}$ is the outputs of the last time step of ${\rm \textbf{H}}^c$.}

For all top-$k$ articles, we integrate all local evidence fragments by concatenation operation.
\begin{equation}\label{eq17}
\setlength{\abovedisplayskip}{2pt}
\setlength{\belowdisplayskip}{2pt}
{\rm \textbf{I}} = [{\rm \textbf{I}}_1; {\rm \textbf{I}}_2; ...; {\rm \textbf{I}}_k]
\end{equation}

\subsection{Dual-View Classification}
\textls[-10]{To alleviate the bias of individual cognition-view evidence fragments and strengthen the consistent shared evidence between global and local evidence, we introduce an inconsistency loss to penalize the disagreement between the both evidence. We define the inconsistency loss function as the Kulllback-Leibler (KL) divergence between ${\rm \textbf{G}}$ and ${\rm \textbf{I}}$.}
\begin{equation}\label{eq18}
\setlength{\abovedisplayskip}{2pt}
\setlength{\belowdisplayskip}{2pt}
{\rm Loss}_{in}={\rm D_{KL}} ({\rm \textbf{G}}||{\rm \textbf{I}})=\sum_{k=1}^{K}{\rm \textbf{G}}_k^{'} {\rm log} \frac{{\rm \textbf{G}}_k^{'}}{{\rm \textbf{I}}_k^{'}}
\end{equation}
where ${\rm \textbf{G}}_k^{'}$ is the $k$-th element of the concatenation of the words in ${\rm \textbf{G}}$, and ${\rm \textbf{I}}_k^{'}$ is the $k$-th element of ${\rm \textbf{I}}$.

\textls[-20]{Furthermore, we fuse the two types of penalized evidence, and adopt softmax function to emit the probability distribution for training, where a loss forces the model to minimize the cross-entropy error for a training sample with ground-truth label $y$:}
\begin{align}
{\rm Loss} &= -\sum y{\rm log}p  \label{eq20} \\
p &= {\rm softmax}({\rm \textbf{W}}_p [{\rm \textbf{G}}; {\rm \textbf{I}}]+{\rm \textbf{b}}_p)  \label{eq20}
\end{align}
where ${\rm \textbf{W}}_p$ and ${\rm \textbf{b}}_p$ are the learnable parameters.

To ensure the effective synergy of the two cognition views, we put together all loss mentioned above for joint training.
\begin{equation}\label{eq22}
\setlength{\abovedisplayskip}{2pt}
\setlength{\belowdisplayskip}{2pt}
\mathcal{L}={\rm Loss}+ \alpha {\rm Loss}_{in}
\end{equation}
where $\alpha$ is the hyper-parameter.

\begin{table*}
\small
	\centering
 \setlength{\tabcolsep}{1.6mm}{
  \begin{tabular}{|l||l|c|c|c|c|c|c|c|c|c|c|c|c|}
		\hline
		\multirow{4}*{Methods} & \multicolumn{8}{c|}{\textbf{Snopes}} & \multicolumn{5}{c|}{\textbf{PolitiFact}} \\
 \cline{2-9} \cline{10-14}
        & & &\multicolumn{3}{c|}{True}&\multicolumn{3}{c|}{False}& & & True & False & Mixed \\ \cline{4-6}
        \cline{7-9} \cline{12-14}
        & micF1 & macF1 & Prec. & Rec. & F1 & Prec. & Rec. & F1 & micF1 & macF1 & F1 & F1 & F1 \\
        \hline \hline
        DeClarE & 0.762 & 0.695 & 0.559 & 0.556 & 0.553 & 0.839 & 0.837 & 0.837 & 0.475 & 0.443 & 0.447 & 0.576 & 0.307 \\
        BERT & 0.781 & 0.706 & 0.587 & 0.601 & 0.594 & 0.852 & 0.854 & 0.853 & 0.493 & 0.462 & 0.478 & 0.596 & 0.320 \\
        HAN & 0.807 & 0.759 & \textbf{0.637} & 0.665 & 0.651 & 0.874 & 0.860 & 0.867 & 0.523 & 0.487 & 0.495 & 0.627 & 0.340 \\
        HAN-ba & 0.771 & 0.738 & 0.556 & 0.765 & 0.644 & \textbf{0.899} & 0.774 & 0.832 & 0.520 & 0.471 & 0.475 & 0.629 & 0.308 \\
        EHIAN & 0.831 & 0.784 & 0.614 & 0.790 & 0.691 & 0.893 & 0.896 & 0.894 & 0.554 & 0.509 & 0.513 & 0.651 & 0.362 \\
        \hline
        CICD (Ours) & \textbf{0.846} & \textbf{0.795} & 0.629 & \textbf{0.796} & \textbf{0.703} & 0.897 & \textbf{0.904} & \textbf{0.900} & \textbf{0.572} & \textbf{0.525} & \textbf{0.529} & \textbf{0.665} & \textbf{0.373} \\
\hline
	\end{tabular}
	\caption{Comparison of our model with baselines on Snopes and PolitiFact.}
\label{Tab1performEval}
}
\end{table*}

\section{Experiments}
\subsection{Datasets and Evaluation Metrics}
\textls[-20]{For evaluation, we utilize three publicly available datasets, i.e., Snopes, PolitiFact (both released by \cite{popat2018declare}), and FEVER \cite{thorne2018fever}. The first two datasets contain 4,341 and 3,568 news claims, associating with 29,242 and 29,556 relevant articles (these articles can be regarded as responses of different individual users to claims) collected from various web sources respectively. FEVER consists of 185,445 claims accompanied by manual annotation Wikipedia articles. For labels, each claim in Snopes is labeled as $true$ and $false$, while PolitiFact divides claims into six kinds of credibility labels: true, mostly true, half true, mostly false, false, and pants on fire. To distinguish the veracity more practically, like Ma et al. \shortcite{ma2019sentence}, we merge mostly true, half true and mostly false into mixed, and treat false and pants on fire as false. Then, the labels of PolitiFact are classified as $true$, $mixed$, and $false$. On FEVER, each claim is partitioned as $supported$, $refuted$, or $N\!EI$ (not enough information). For evaluation metrics, on Snopes and PolitiFact, we exploit micro-/macro-averaged F1(micF1/macF1), class-specific precision (Prec.), recall (Rec.) and F1-score (F1) as evaluation metrics. We hold out 10\% of the claims for tuning the hyper parameters, and conduct 5-fold cross-validation on the rest of the claims. On FEVER, we leverage accuracy (Acc.), and F1-score (F1) as evaluation metrics, and follow Thorne et al. \shortcite{thorne2018fever} to partition the annotated claims into training, development (Dev.), and testing (Test.) sets.}

\vspace{-0.3em}
\subsection{Settings}
\textls[-20]{For parameter configurations, we adjust them according to the performance of development sets, we set the word embedding size $d$ to 768. The dimensionality of LSTM hidden states $d_h$ is 120. The length $l$ of each relevant article is 100 and that of the claim $p$ is assigned as 20. Due to no parameters depend on the number of articles $N$, instead of intercepting a fixed number, we set $N$ to vary with claims. Initial learning rate is set to 2e-3. The loss weight coefficient $\alpha$ is trained to 0.2. The dropout rate is 0.4, and we set the mini-batch size of the three datasets as 32, 32, and 64, respectively. Additionally, an Adam \cite{kingma2014adam} optimizer with $\beta1$ as 0.9 and $\beta2$ as 0.999 is used to optimize all trainable parameters.}

\vspace{-0.3em}
\subsection{Experiments on Snopes and PolitiFact}
\subsubsection{Performance Comparison}
\label{sec431perfm}
\textls[-25]{We compare CICD and several competitive baselines: 1) \textbf{DeClarE} \cite{popat2018declare} models joint interactions between claims and articles and aggregates word-level credibility signals from external articles for evidence-aware assessment; 2) \textbf{BERT} \cite{devlin2019bert}, we employ pre-trained BERT classifier to verify claims; 3) \textbf{HAN} \cite{ma2019sentence}, a hierarchical attention network, constructs the interactions between claims and relevant articles for capturing sentence-level evidence by considering their topical coherence and semantic inference strength; 4) \textbf{HAN-ba} \cite{ma2019sentence} is a variant of HAN, where the gated attention is replaced to biaffine attention for acquiring evidence; and 5) \textbf{EHIAN} \cite{wu2020evidence} is an evidence-aware hierarchical interactive attention network, which focuses on the direct interaction between claim and relevant articles to explore key evidence fragments. As shown in Table \ref{Tab1performEval}, we observe that:}

%

\begin{itemize}[leftmargin=*]
\item \textls[-15]{BERT achieves at least 6.5\% improvement on micF1 than DeClarE, which illustrates pre-trained model can learn rich semantic context features to improve performance, which is also the reason that we adopt BERT to train word embeddings. HAN consistently outperforms BERT, which indicates HAN capturing the coherence between relevant articles could help improve the task performance.}
\item \textls[-20]{In interpretable methods, CICD outperforms DeClarE, which is because our model not only focuses on word-level semantics like DeClarE, but also grasps the holistic sentence-level features. Moreover, owing to HAN and HAN-ba drive all relevant articles to participate in the interaction, prompting them to gain a small boost in precision on Snopes, but this way may introduce noise from nonsignificant articles. CICD effectively avoids this problem by selecting vital articles for interaction, which obtains significant improvements in other metrics compared with HAN and HAN-ba. Furthermore, CICD consistently outperforms EHIAN on Snopes and PolitiFact. The superiority is clear: CICD not only values individual cognition view to capture key evidence fragments, but also generates collective cognition-view evidence for claim verification.}
\end{itemize}

\begin{table}
\small
	\centering
\setlength{\tabcolsep}{1.8mm}{
	\begin{tabular}{|r||l|c|c|c|}
		\hline
		\multirow{2}*{Methods} & \multicolumn{2}{c|}{Snopes} & \multicolumn{2}{c|}{PolitiFact} \\ \cline{2-3} \cline{4-5}
        & micF1 & macF1 & micF1 & macF1 \\  \hline \hline
        -matching U & 0.802 & 0.753 & 0.529 & 0.486 \\
        -CED & 0.791 & 0.748 & 0.526 & 0.476 \\
        -selected I & 0.810 & 0.763 & 0.541 & 0.490 \\
        -interaction I & 0.822 & 0.770 & 0.557 & 0.497 \\
        -ISI & 0.803 & 0.751 & 0.530 & 0.483 \\
        -inconsistency loss & 0.831 & 0.782 & 0.556 & 0.508\\
        CICD & 0.846 & 0.795 & 0.572 & 0.525 \\  \hline
	\end{tabular}
	\caption{Ablation analysis of CICD.}
\label{Tab2ablation}
}
\end{table}

\subsubsection{Ablation Study}
\textls[-20]{In order to evaluate the impact of each component of CICD, we ablate CICD into the following simplified models: 1) \textbf{-matching U} represents the attention-based matching layer of CED is removed; 2) \textbf{-CED} means CED is deleted from our model; 3) \textbf{-selected I} refers to the selected mechanism is removed from ISI; 4) \textbf{-interaction I} represents the co-interaction unit of ISI is replaced by concatenation operation; 5) \textbf{-ISI} corresponds to ISI is separated; and 6) \textbf{-inconsistency loss} means the inconsistency loss is removed.
As shown in Table \ref{Tab2ablation}, we observe that:}
\begin{itemize}[leftmargin=*]
\item \textls[-25]{The removal of each module (-CED or -ISI) weakens the performance of CICD, presenting from 4.2\% to 5.5\% degradation in micF1, and the stripping of different layers (like -selected I and -interaction I) of each module also reduces the model performance, reducing at least 2.4\% performance in micF1, which describes the effectiveness of each component and the organic integrity of CICD.}
\item \textls[-20]{-CED reflects the lowest performance in all simplified models, decreasing 5.5\% and 4.6\% in micF1 on the two datasets, respectively, which elaborates the effectiveness of our CICD capturing the collective cognition-view global evidence. Meanwhile, -ISI underperforms CICD, showing 4.3\% and 4.2\% degradation in micF1 on the two datasets respectively, which conveys the necessity of the exploration of local key evidence fragments from individual cognition view.}
\item \textls[-20]{When compared with -inconsistency loss, CICD significantly improves the performance on the two datasets with the help of inconsistency loss unit, which verifies the effectiveness of our model relying on inconsistency loss to discover shared valuable semantics between global and local evidence.}
\end{itemize}

\begin{figure*}
\centering
\subfigure[On Snopes]{
    \begin{minipage}[b]{0.48\textwidth}
    \includegraphics[width=1\textwidth]{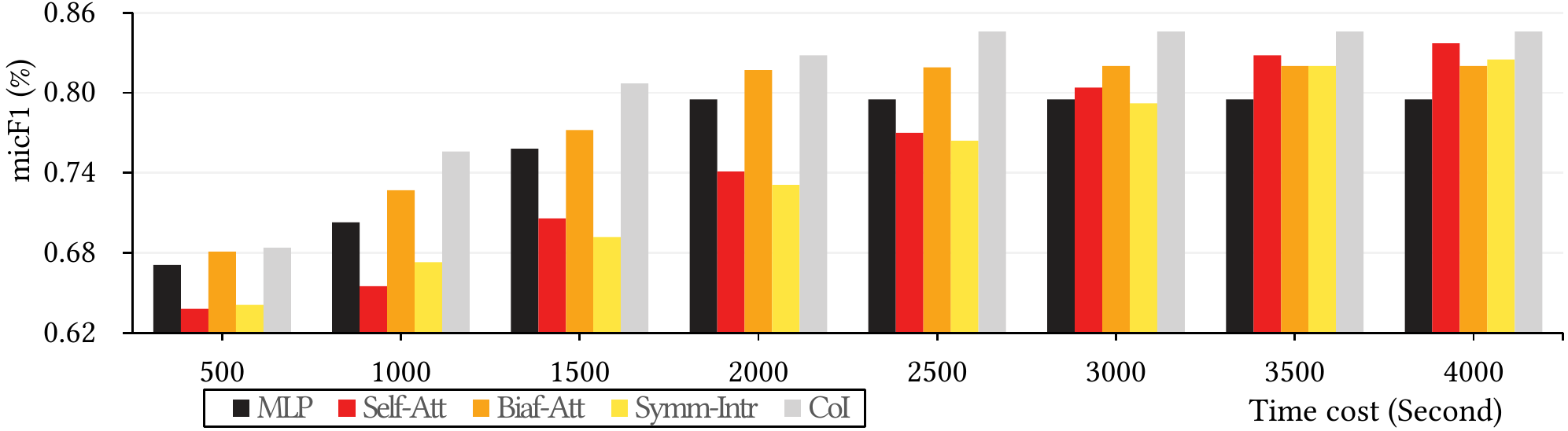}
    \end{minipage}
}
\subfigure[On PolitiFact]{
    \begin{minipage}[b]{0.48\textwidth}
    \includegraphics[width=1\textwidth]{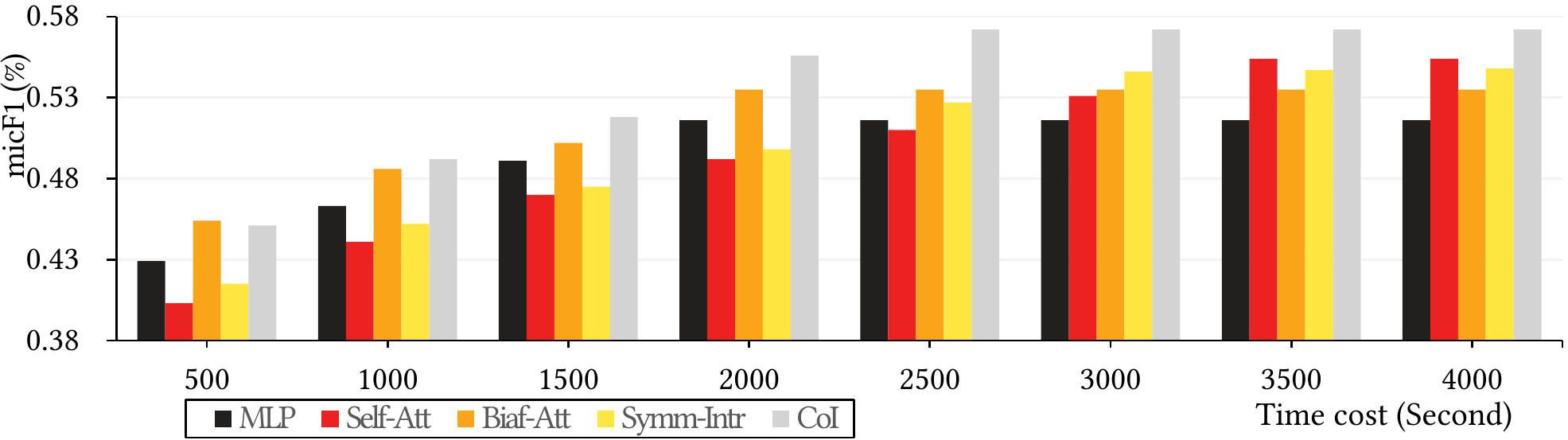}
    \end{minipage}
}
\caption{\textls[-30]{The performance comparison between co-interaction networks (CoI) and some prevalent interaction networks.}}
\label{fig3interaction}
\end{figure*}

\vspace{-0.3em}
\subsubsection{Evaluation of Co-Interaction Networks}
\textls[-15]{To obtain a more detailed understanding of the superiority of our co-interaction networks (CoI), we compare CoI with the following prevalent interaction networks: 1) \textbf{MLP} (\textit{Multilayer Perceptron}) acts as an interaction strategy to automatically abstract the integrated representation of claims and articles; 2) \textbf{Self-Att} (\textit{Self-attention Networks}) \cite{vaswani2017attention} adopts the claim as query, and relevant articles to serve as values and keys for interaction; 3) \textbf{Biaf-Att} (\textit{Biaffine Attention}) \cite{ma2019sentence} measures the degree of semantic matching for interaction; and 4) \textbf{Symm-Intr} (\textit{Symmetric interaction attention}) \cite{tao2019one} is exploited to model the interaction between claims and articles. Specifically, we investigate the performance and time cost of these methods on Snopes and PolitiFact based on Linux CentOS with NVIDIA TITAN Xp GPU, as shown in Figure \ref{fig3interaction}. We observe that:}

\textls[-15]{From the overall performance of all methods, our method achieves the optimal performance, outperforming other methods by more than 5.1\% and 5.6\% performance in micF1, respectively. From the indicator of time cost, our method saves a great deal of time. Compared with Self-Att and Symm-Intr, our method saves from 500 to 1,000 seconds in time cost on the two datasets, respectively. The reason is that the structures of multiple mappings of self-attention networks and the repeat stacks of symmetric attention delay the efficiency. Although the time cost of our method is higher than that of MLP and Biaf-Att, the performance of both methods is unsatisfactory, which is lower than our method al least 2.6\% and 3.7\% in micF1 on both datasets. On the whole, these adequately manifest the superiority of our method.}

\vspace{-0.2em}
\subsubsection{\textls[-10]{Evaluation of Hierarchical Attention Decoder}}
\textls[-15]{To verify the effectiveness of the internal structure of hierarchical attention decoder (HAD) in CED, we ablate HAD with the following models: \textbf{-word.}, \textbf{-sentence.}, and \textbf{-merge.} respectively denote HAD removing word-level attention $\alpha$, sentence-level attention $\beta$, and merged semantics $\gamma$. \textbf{decoder.} represents the vanilla decoder. Experimental results are shown in Table \ref{Tab3evalDecoder}, we observe that: first, the removal of any module of HAD could weaken the performance of the model, which confirms the effectiveness of each module. Second, in addition to the basic decoder, our model achieves the most prominent boost with the support of sentence-level attention, which proves the effectiveness of HAD fusing sentence-level semantics to capture global semantics of HAD.}
\begin{table}
\small
	\centering
	\begin{tabular}{|r||l|c|c|c|}
    \hline
		\multirow{2}*{Methods} & \multicolumn{2}{c|}{Snopes} & \multicolumn{2}{c|}{PolitiFact} \\ \cline{2-3} \cline{4-5}
        & micF1 & macF1 & micF1 & macF1 \\  \hline \hline
        -word. & 0.827 & 0.784 & 0.558 & 0.517 \\
        -sentence. & 0.815 & 0.770 & 0.545 & 0.502 \\
        -merge. & 0.833 & 0.787 & 0.562 & 0.520 \\
        decoder. & 0.804 & 0.758 & 0.535 & 0.489 \\
        CICD & 0.846 & 0.795 & 0.572 & 0.525 \\ \hline
	\end{tabular}
	\caption{Evaluation of hierarchical attention decoder.}
\label{Tab3evalDecoder}
\end{table}

\begin{table}
\small
	\centering
	\begin{tabular}{|l||l|c|c|c|}
		\hline
        \multirow{2}*{Methods} & \multicolumn{2}{c|}{Dev.} & \multicolumn{2}{c|}{Test.} \\  \cline{2-3} \cline{4-5}
		& Acc. & F1 & Acc. & F1 \\ \hline
        NSMN & 0.697 & 0.431 & 0.621 & 0.398 \\
        HAN & 0.720 &	0.488 & 0.669 & 0.446  \\
        GEAR & 0.738 & 0.492 & 0.708 & 0.474 \\
        KGAT & 0.745 & 0.501 & 0.716 & 0.485 \\
        CICD (Ours) & \textbf{0.763} & \textbf{0.525} &	\textbf{0.731} & \textbf{0.497}  \\ \hline
	\end{tabular}
	\caption{Results of different baselines on FEVER.}
	\label{Tab3resultsFever}
\end{table}

\begin{figure}
	\centering
	\includegraphics[width=0.48\textwidth]{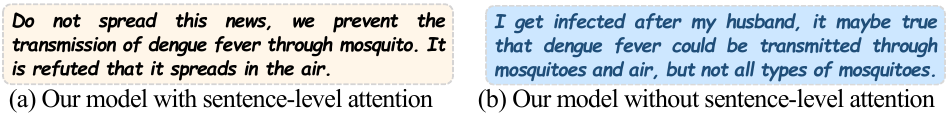}
	\caption{The sequences generated by our model with and without sentence-level attention, respectively.}
	\label{fig4sentence}
\end{figure}

\begin{figure*}
	\centering
	\includegraphics[width=1\textwidth]{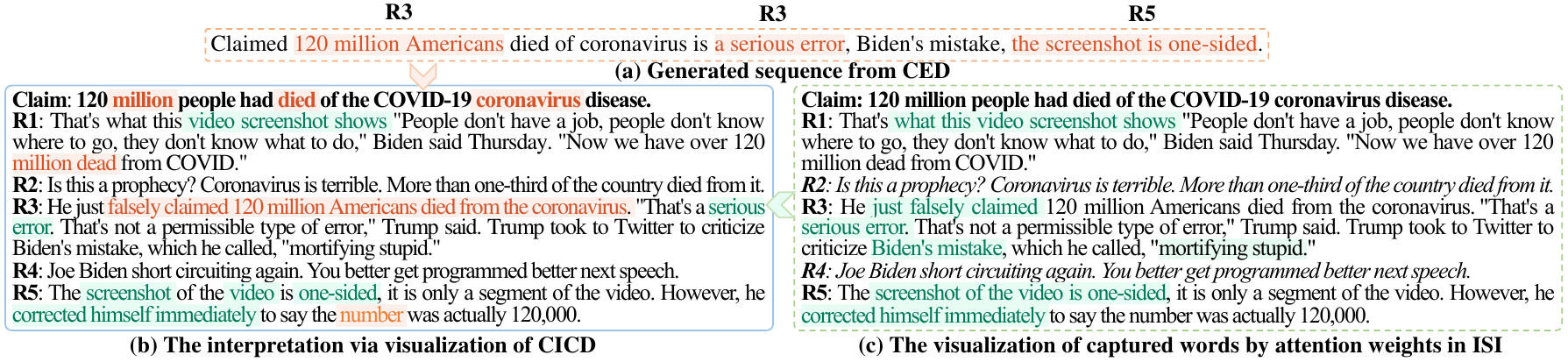}
	\caption{The Interpretability and transparency of different modules of CICD via a false sample on Snopes.}
	\label{fig4visualization}
\end{figure*}

\textls[-20]{To further investigate the contribution of sentence-level semantics to the global evidence, we take Figure \ref{fig1example} as an example to visualize the global evidence generated by our model with and without sentence-level attention, respectively. As shown in Figure \ref{fig4sentence}, we observe that the model with sentence-level attention focuses more on the sentences with maximum weight, that is, R4, such as the words `do not spread' and `refuted it spreads in the air', while the model without sentence-level attention does not identify which relevant articles are more valuable, so that they concentrate more on R2 and R3, like `get infected husband' and `not all types of mosquitoes'. These fully prove the effectiveness of sentence-level semantics for the generation of global evidence.}

\vspace{-0.3em}
\subsection{Experiments on FEVER}
\textls[-20]{To examine the extensibility of our model, we also compare CICD and the following state-of-the-art baselines on FEVER dataset: 1) \textbf{NSMN}: The pipeline-based system, Neural Semantic Matching Network \cite{nie2019combining}, conducts document retrieval, sentence selection, and claim verification jointly for fact extraction and verification; 2) \textbf{HAN}: It has introduced in Section 4.3.1; 3) \textbf{GEAR}: A graph-based evidence aggregating and reasoning model \cite{zhou2019gear} enables information to transfer on a fully-connected evidence graph and then utilizes different aggregators to collect multi-evidence information; 4) \textbf{KGAT}: Kernel graph attention network \cite{liu2020fine} conducts more fine-grained fact verification with kernel-based attentions, where using BERT (Base) encoder with ESIM retrieved sentences.}

\textls[-20]{As shown in Table \ref{Tab3resultsFever}, we observe that: CICD outperforms the two pipelines (NSMN and HAN) by from 4.3\% to 11.0\% boost in accuracy, respectively. This is because these two baselines lack the integration and reasoning process between relevant articles when capturing evidence. CICD boosts the performance in comparison with GEAR and KGAT, showing at least 1.8\% and 1.5\% improvement in accuracy on development and testing sets, respectively. The reason may be that although the two graph-based models aggregate and reason information from relevant articles to collect multi-evidence, they treat each relevant article equally, leading to individual-cognitive relevant articles with some biased semantics interfering with their reasoning process. It is more feasible for our model to discover global evidence and local key evidence fragments comprehensively from the perspectives of collective and individual cognition.}

\subsection{Case Study: Cognition-view Explanation Analysis}
\textls[-20]{To interpret the results of our model more transparently and intuitively, we visualize the outputs of each module of CICD as shown in Figure \ref{fig4visualization}, where Figure \ref{fig4visualization} (a) is the sequence generated by CED module, and the highlighted words in Figure \ref{fig4visualization}(b) and \ref{fig4visualization} (c) are respectively the words captured by CICD to interpret the results and the words obtained by ISI module to obtain the evidence fragments. We could learn:}
\begin{itemize}[leftmargin=*]
\item \textls[-10]{ISI ignores some articles with pale and feeble semantics (R2 and R4), and selects the articles with more valuable semantics (R1, R3, and R5) and captures multiple local evidence fragments, such as `this video screenshot shows' (E1), `serious error' (E2), and `screenshot of the video is one-sided' (E3). Particularly, fragment E1 is misleading, which reflects the deviation of individual cognition.}
\item \textls[-20]{The sequence generated by CED effectively gains available evidence `120 million Americans a serious error' and `the screenshot is one-sided' through balancing the possible evidence semantics in relevant articles from a global perspective.}
\item \textls[-15]{By constraining global and local evidence, CICD disciplines the misleading evidence fragment E1 captured by ISI, and finally highlights the shared salient evidence between the both as the final interpretability of the verification results.}
\end{itemize}

\section{Conclusion}
\textls[-10]{In this paper, we proposed a unified dual-view model based on the perspectives of collective and individual cognition for interpretable claim verification, which constructed collective cognition view-based encoder-decoder module to generate global evidence and designed individual cognition view-based selected interaction module to explore local key evidence segments. Besides, we introduced inconsistent loss to penalize the disagreement between global and local evidence for promoting the capture of consistent shared evidence. Experiments on three different widely used datasets demonstrated the effectiveness and interpretability of our model. In the future, we plan to expand the work as follows: 1) Developing questioning mechanism to filter the suspicious evidence; and 2) Integrating social cognition, psychology, and other interdisciplinary knowledge to improve the interpretability of claim verification.}

\section*{Acknowledgments}
The research work was supported by National Key Research and Development Program in China (2019YFB2102300); The World-Class Universities (Disciplines) and the Characteristic Development Guidance Funds for the Central Universities of China (PY3A022); Ministry of Education Fund Projects$\!$ (18JZD022 and 2017B00030); Shenzhen Science and Technology Project (JCYJ20180306170836595); Basic Scientific Research Operating Expenses of Central Universities (ZDYF2017006); Xi'an Navinfo Corp.\& Engineering Center of Xi'an Intelligence Spatial-temporal Data Analysis Project (C2020103); Beilin District of Xi'an Science \& Technology Project (GX1803). We would like to thank the anonymous reviewers for their valuable and constructive comments.

\bibliographystyle{acl_natbib}
\bibliography{acl2021_refer}

\begin{thebibliography}{33}
\expandafter\ifx\csname natexlab\endcsname\relax\def\natexlab#1{#1}\fi

\bibitem[{Ajao et~al.(2019)Ajao, Bhowmik, and Zargari}]{ajao2019sentiment}
Oluwaseun Ajao, Deepayan Bhowmik, and Shahrzad Zargari. 2019.
\newblock Sentiment aware fake news detection on online social networks.
\newblock In \emph{ICASSP 2019-2019 IEEE International Conference on Acoustics,
  Speech and Signal Processing (ICASSP)}, pages 2507--2511. IEEE.

\bibitem[{Allen et~al.(2020)Allen, Howland, Mobius, Rothschild, and
  Watts}]{allen2020evaluating}
Jennifer Allen, Baird Howland, Markus Mobius, David Rothschild, and Duncan~J
  Watts. 2020.
\newblock Evaluating the fake news problem at the scale of the information
  ecosystem.
\newblock \emph{Science Advances}, 6(14):eaay3539.

\bibitem[{Boogert et~al.(2018)Boogert, Madden, Morand-Ferron, and
  Thornton}]{Boogert2018MeasuringAU}
N.~Boogert, J.~Madden, J.~Morand-Ferron, and A.~Thornton. 2018.
\newblock Measuring and understanding individual differences in cognition.
\newblock \emph{Philosophical Transactions of the Royal Society B: Biological
  Sciences}, 373.

\bibitem[{Bovet and Makse(2019)}]{bovet2019influence}
Alexandre Bovet and Hern{\'a}n~A Makse. 2019.
\newblock Influence of fake news in twitter during the 2016 us presidential
  election.
\newblock \emph{Nature communications}, 10(1):1--14.

\bibitem[{Devlin et~al.(2019)Devlin, Chang, Lee, and
  Toutanova}]{devlin2019bert}
Jacob Devlin, Ming-Wei Chang, Kenton Lee, and Kristina Toutanova. 2019.
\newblock Bert: Pre-training of deep bidirectional transformers for language
  understanding.
\newblock In \emph{Proceedings of the 2019 Conference of the North American
  Chapter of the Association for Computational Linguistics: Human Language
  Technologies, Volume 1 (Long and Short Papers)}, pages 4171--4186.

\bibitem[{Fleming(2020)}]{nic2020coronavirus}
Nic Fleming. 2020.
\newblock Coronavirus misinformation, and how scientists can help to fight it.
\newblock \emph{Nature}, 583:155--156.

\bibitem[{Goldstein and Naglieri(2011)}]{Goldstein2011halo}
Sam Goldstein and Jack~A. Naglieri, editors. 2011.
\newblock \emph{Halo Effect}, pages 725--725. Springer US, Boston, MA.

\bibitem[{Greenwald et~al.(1998)Greenwald, McGhee, and
  Schwartz}]{Greenwald1998MeasuringID}
A.~Greenwald, D.~McGhee, and J.~L. Schwartz. 1998.
\newblock Measuring individual differences in implicit cognition: the implicit
  association test.
\newblock \emph{Journal of personality and social psychology}, 74 6:1464--80.

\bibitem[{Hoffman(2018)}]{Hoffman2018AnID}
Paul Hoffman. 2018.
\newblock An individual differences approach to semantic cognition: Divergent
  effects of age on representation, retrieval and selection.
\newblock \emph{Scientific Reports}, 8.

\bibitem[{Ji et~al.(2019)Ji, Holmes, MacLeod, and Murphy}]{Ji2019SpontaneousCI}
J.~Ji, E.~Holmes, C.~MacLeod, and F.~Murphy. 2019.
\newblock Spontaneous cognition in dysphoria: reduced positive bias in
  imagining the future.
\newblock \emph{Psychological Research}, 83:817 -- 831.

\bibitem[{Kingma and Ba(2015)}]{kingma2014adam}
Diederik~P Kingma and Jimmy Ba. 2015.
\newblock Adam: A method for stochastic optimization.
\newblock In \emph{ICLR (Poster)}.

\bibitem[{Kumar and Carley(2019)}]{kumar2019tree}
Sumeet Kumar and Kathleen~M Carley. 2019.
\newblock Tree lstms with convolution units to predict stance and rumor
  veracity in social media conversations.
\newblock In \emph{Proceedings of the 57th Annual Meeting of the Association
  for Computational Linguistics}, pages 5047--5058.

\bibitem[{Liu et~al.(2019)Liu, Liu, and Ren}]{liu2019trust}
Shuaipeng Liu, Shuo Liu, and Lei Ren. 2019.
\newblock Trust or suspect? an empirical ensemble framework for fake news
  classification.
\newblock In \emph{Proceedings of the 12th ACM International Conference on Web
  Search and Data Mining}, pages 11--15.

\bibitem[{Liu et~al.(2020)Liu, Xiong, Sun, and Liu}]{liu2020fine}
Zhenghao Liu, Chenyan Xiong, Maosong Sun, and Zhiyuan Liu. 2020.
\newblock Fine-grained fact verification with kernel graph attention network.
\newblock In \emph{Proceedings of the 58th Annual Meeting of the Association
  for Computational Linguistics}, pages 7342--7351.

\bibitem[{Ma et~al.(2019)Ma, Gao, Joty, and Wong}]{ma2019sentence}
Jing Ma, Wei Gao, Shafiq Joty, and Kam-Fai Wong. 2019.
\newblock Sentence-level evidence embedding for claim verification with
  hierarchical attention networks.
\newblock In \emph{ACL}, pages 2561--2571.

\bibitem[{Nguyen et~al.(2020)Nguyen, Sugiyama, Nakov, and Kan}]{nguyen2020fang}
Van-Hoang Nguyen, Kazunari Sugiyama, Preslav Nakov, and Min-Yen Kan. 2020.
\newblock Fang: Leveraging social context for fake news detection using graph
  representation.
\newblock In \emph{Proceedings of the 29th ACM International Conference on
  Information \& Knowledge Management}, pages 1165--1174.

\bibitem[{Nie et~al.(2019)Nie, Chen, and Bansal}]{nie2019combining}
Yixin Nie, Haonan Chen, and Mohit Bansal. 2019.
\newblock Combining fact extraction and verification with neural semantic
  matching networks.
\newblock In \emph{Proceedings of the AAAI Conference on Artificial
  Intelligence}, volume~33, pages 6859--6866.

\bibitem[{Popat et~al.(2018)Popat, Mukherjee, Yates, and
  Weikum}]{popat2018declare}
Kashyap Popat, Subhabrata Mukherjee, Andrew Yates, and Gerhard Weikum. 2018.
\newblock Declare: Debunking fake news and false claims using evidence-aware
  deep learning.
\newblock In \emph{Proceedings of the 2018 Conference on Empirical Methods in
  Natural Language Processing}, pages 22--32.

\bibitem[{Przybyla(2020)}]{przybyla2020capturing}
Piotr Przybyla. 2020.
\newblock Capturing the style of fake news.
\newblock In \emph{Proceedings of the AAAI Conference on Artificial
  Intelligence}, volume~34, pages 490--497.

\bibitem[{Shu et~al.(2019)Shu, Zhou, Wang, Zafarani, and Liu}]{shu2019role}
Kai Shu, Xinyi Zhou, Suhang Wang, Reza Zafarani, and Huan Liu. 2019.
\newblock The role of user profiles for fake news detection.
\newblock In \emph{Proceedings of the 2019 IEEE/ACM International Conference on
  Advances in Social Networks Analysis and Mining}, pages 436--439.

\bibitem[{Tao et~al.(2019)Tao, Wu, Xu, Hu, Zhao, and Yan}]{tao2019one}
Chongyang Tao, Wei Wu, Can Xu, Wenpeng Hu, Dongyan Zhao, and Rui Yan. 2019.
\newblock One time of interaction may not be enough: Go deep with an
  interaction-over-interaction network for response selection in dialogues.
\newblock In \emph{Proceedings of the 57th Annual Meeting of the Association
  for Computational Linguistics}, pages 1--11.

\bibitem[{Thorne et~al.(2018)Thorne, Vlachos, Christodoulopoulos, and
  Mittal}]{thorne2018fever}
James Thorne, Andreas Vlachos, Christos Christodoulopoulos, and Arpit Mittal.
  2018.
\newblock Fever: a large-scale dataset for fact extraction and verification.
\newblock In \emph{Proceedings of the 2018 Conference of the North American
  Chapter of the Association for Computational Linguistics: Human Language
  Technologies, Volume 1 (Long Papers)}, pages 809--819.

\bibitem[{Troyer(2011)}]{Troyer2011}
Angela~K. Troyer. 2011.
\newblock \emph{Primacy Effect}, pages 2017--2018. New York, NY.

\bibitem[{Vaswani et~al.(2017)Vaswani, Shazeer, Parmar, Uszkoreit, Jones,
  Gomez, Kaiser, and Polosukhin}]{vaswani2017attention}
Ashish Vaswani, Noam Shazeer, Niki Parmar, Jakob Uszkoreit, Llion Jones,
  Aidan~N Gomez, {\L}ukasz Kaiser, and Illia Polosukhin. 2017.
\newblock Attention is all you need.
\newblock In \emph{Advances in neural information processing systems}, pages
  5998--6008.

\bibitem[{Willard and Norenzayan(2017)}]{Willard2017SpiritualBN}
Aiyana~K. Willard and A.~Norenzayan. 2017.
\newblock “spiritual but not religious”: Cognition, schizotypy, and
  conversion in alternative beliefs.
\newblock \emph{Cognition}, 165:137--146.

\bibitem[{Wu et~al.(2019)Wu, Rao, Jin, Nazir, and Sun}]{wu2019different}
Lianwei Wu, Yuan Rao, Haolin Jin, Ambreen Nazir, and Ling Sun. 2019.
\newblock Different absorption from the same sharing: Sifted multi-task
  learning for fake news detection.
\newblock In \emph{Proceedings of the 2019 Conference on Empirical Methods in
  Natural Language Processing and the 9th International Joint Conference on
  Natural Language Processing (EMNLP-IJCNLP)}, pages 4636--4645.

\bibitem[{Wu et~al.(2020{\natexlab{a}})Wu, Rao, Liang, Nazir
  et~al.}]{wu2020dtca}
Lianwei Wu, Yuan Rao, Hao Liang, Ambreen Nazir, et~al. 2020{\natexlab{a}}.
\newblock Dtca: Decision tree-based co-attention networks for explainable claim
  verification.
\newblock In \emph{Proceedings of the 58th Annual Meeting of the Association
  for Computational Linguistics}, pages 1024--1035.

\bibitem[{Wu et~al.(2020{\natexlab{b}})Wu, Rao, Nazir, and
  Jin}]{wu2020discovering}
Lianwei Wu, Yuan Rao, Ambreen Nazir, and Haolin Jin. 2020{\natexlab{b}}.
\newblock Discovering differential features: Adversarial learning for
  information credibility evaluation.
\newblock \emph{Information Sciences}, 516:453--473.

\bibitem[{Wu et~al.(2021)Wu, Rao, Sun, and He}]{wu2021evidence}
Lianwei Wu, Yuan Rao, Ling Sun, and Wangbo He. 2021.
\newblock Evidence inference networks for interpretable claim verification.
\newblock In \emph{Proceedings of the AAAI Conference on Artificial
  Intelligence}, pages 1165--1174.

\bibitem[{Wu et~al.(2020{\natexlab{c}})Wu, Rao, Yang, Wang, and
  Nazir}]{wu2020evidence}
Lianwei Wu, Yuan Rao, Xiong Yang, Wanzhen Wang, and Ambreen Nazir.
  2020{\natexlab{c}}.
\newblock Evidence-aware hierarchical interactive attention networks for
  explainable claim verification.
\newblock In \emph{Proceedings of IJCAI}.

\bibitem[{Yang et~al.(2019)Yang, Shu, Wang, Gu, Wu, and
  Liu}]{yang2019unsupervised}
Shuo Yang, Kai Shu, Suhang Wang, Renjie Gu, Fan Wu, and Huan Liu. 2019.
\newblock Unsupervised fake news detection on social media: A generative
  approach.
\newblock In \emph{Proceedings of the AAAI Conference on Artificial
  Intelligence}, volume~33, pages 5644--5651.

\bibitem[{Zhou et~al.(2019)Zhou, Han, Yang, Liu, Wang, Li, and
  Sun}]{zhou2019gear}
Jie Zhou, Xu~Han, Cheng Yang, Zhiyuan Liu, Lifeng Wang, Changcheng Li, and
  Maosong Sun. 2019.
\newblock Gear: Graph-based evidence aggregating and reasoning for fact
  verification.
\newblock In \emph{Proceedings of the 57th Annual Meeting of the Association
  for Computational Linguistics}, pages 892--901.

\bibitem[{Zhou et~al.(2020)Zhou, Wu, and Zafarani}]{zhou2020safe}
Xinyi Zhou, Jindi Wu, and Reza Zafarani. 2020.
\newblock Safe: Similarity-aware multi-modal fake news detection.
\newblock In \emph{Pacific-Asia Conference on Knowledge Discovery and Data
  Mining}, pages 354--367. Springer.

\end{thebibliography}

\end{document}